\documentclass[letterpaper, 10 pt, conference]{ieeeconf}  

\usepackage{hyperref}
\usepackage{url}
\usepackage{graphicx} 
\usepackage{booktabs}
\usepackage{multirow}
\usepackage{subfigure}
\usepackage{algorithm}
\usepackage{algpseudocode}
\usepackage{caption}
\usepackage{cite}
\usepackage{xcolor}
\usepackage{amssymb}
\usepackage{amsmath}

\definecolor{codegreen}{rgb}{0,0.6,0}
\definecolor{codegray}{rgb}{0.5,0.5,0.5}
\definecolor{codepurple}{rgb}{0.58,0,0.82}
\definecolor{backcolour}{rgb}{0.95,0.95,0.92}

\IEEEoverridecommandlockouts                              

\title{\LARGE \bf
Topo-Field: Topometric mapping with Brain-inspired Hierarchical Layout-Object-Position Fields
}

\author{Jiawei Hou$^{1}$, Wenhao Guan$^{1}$, Longfei Liang$^{2}$, Jianfeng Feng$^{3}$, Xiangyang Xue$^{1}$, and Taiping Zeng$^{3,*}$
\thanks{*Corresponding author}
\thanks{$^{1}$ School of Computer Science, Fudan University, Shanghai, China
        {\tt\small \{jwhou23, whguan21\}@m.fudan.edu.cn xyxue@fudan.edu.cn}}%
\thanks{$^{2}$ Shanghai NeuHelium Neuromorphic Intelligence Tech. Co., Ltd., Shanghai, China
        {\tt\small longfei.liang@neuhelium.com}}%
\thanks{$^{3}$ Institute of Science and Technology for Brain-Inspired Intelligence, Fudan University, Shanghai, China
        {\tt\small \{jffeng, zengtaiping\}@fudan.edu.cn}}%
}

\begin{document}

\maketitle
\thispagestyle{empty}
\pagestyle{empty}

\begin{abstract}

Mobile robots require comprehensive scene understanding to operate effectively in diverse environments, enriched with contextual information such as layouts, objects, and their relationships.
Although advances like neural radiation fields (NeRFs) offer high-fidelity 3D reconstructions, they are computationally intensive and often lack efficient representations of traversable spaces essential for planning and navigation. In contrast, topological maps are computationally efficient but lack the semantic richness necessary for a more complete understanding of the environment.
Inspired by a population code in the postrhinal cortex (POR) strongly tuned to spatial layouts over scene content rapidly forming a high-level cognitive map, this work introduces Topo-Field, a framework that integrates Layout-Object-Position (LOP) associations into a neural field and constructs a topometric map from this learned representation.
LOP associations are modeled by explicitly encoding object and layout information, while a Large Foundation Model (LFM) technique allows for efficient training without extensive annotations. The topometric map is then constructed by querying the learned neural representation, offering both semantic richness and computational efficiency.
Empirical evaluations in multi-room environments demonstrate the effectiveness of Topo-Field in tasks such as position attribute inference, query localization, and topometric planning, successfully bridging the gap between high-fidelity scene understanding and efficient robotic navigation. See the project website at \href{https://jarvishou829.github.io/Topo-Field/}{https://jarvishou829.github.io/Topo-Field/}.

\end{abstract}

\section{Introduction}\label{introduction}

Mobile robots are rapidly moving from research labs to widespread use. For these robots to operate autonomously in complex environments, a deep understanding of their surroundings is crucial~\cite{cadena2016past}. 
Hierarchical graph-like scene representation along with detailed environmental reconstruction enabling efficient path planning, will be key for robotic deployment in real-world scenarios~\cite{blochliger2018topomap}.

Recently, detailed environmental reconstruction has made great progress in producing lifelike 3D reconstructions~\cite{sfm,svo,bundlefusion,banet}, in which NeRF~\cite{nerf} is a prime instance. As improvements, works like~\cite{snerf-1, snerf-2, snerf-3} introduce semantic information for better scene understanding. Further, features powered by Large-Foundation-Models (LFMs), trained on massive datasets across various scenes, are employed with general knowledge for open scene understanding~\cite{clip-fields, vlmaps, lerf2023}. However, it is computationally demanding and lacks global layout information using detailed neural fields for planning and navigation.

In contrast, existing topological maps for path planning and navigation in complex environments are often derived from LiDAR Simultaneous-Localization-and-Mapping (SLAM) using 3D dense submaps~\cite{gomez2020hybrid} or visual SLAM by clustering free-space regions and extracting occupancy information from point clouds~\cite{blochliger2018topomap}. While this approach increases path planning accuracy, computing topology with traditional methods comes with high computational costs and tends to strip away essential semantic information, reducing the robot's ability to fully understand and interpret the environment, which is critical for advanced autonomous functions such as language/image-prompted localization and navigation.

\begin{figure}[t]
\centering
\includegraphics[width=\linewidth]{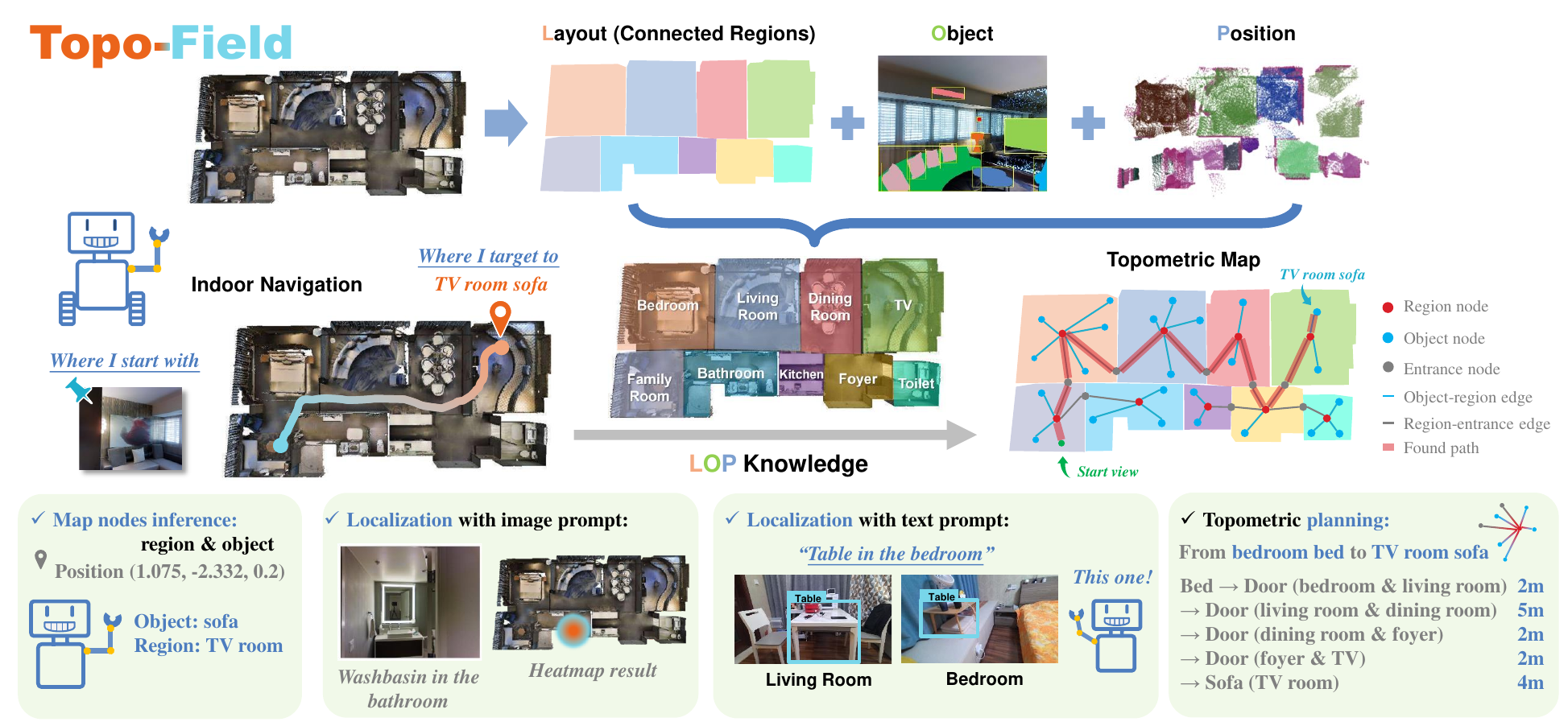}
\vspace{-0.1in}
\caption{\textbf{Illustration of the Topo-Field strategy and capabilities.} Hierarchically dividing scene information into layout, object, and position to model them explicitly, layout-object-position associated knowledge enables robots with a topometric map representing the scene and planning navigable path to realize a more comprehensive spatial cognition.}
\label{fig:idea}
\vspace{-0.15in}
\end{figure}

To this end, we propose to build a neural representation as spatial knowledge and construct a topometric map based on this, originating from a brain-inspired approach. Theoretically, neuroscientists have long discovered that animals process their surroundings using topological coding, forming what is known as a "cognitive map"~\cite{tolman1948cognitive}, a concept embodied by place cells~\cite{o1971hippocampus}. These place cells, along with spatial view cells~\cite{rolls1998information}, respond to specific scene contents. More recently, research has shown that a population code in the postrhinal cortex (POR) is strongly tuned to spatial layout rather than scene content~\cite{por-cells}, capturing spatial representations relative to environmental centers to form a high-level cognitive map from egocentric perception to allocentric understanding~\cite{zeng2022theory}.

Most related works either do not explicitly represent the layout features~\cite{yang2023regionplc} or build the topo-map in a clustering and incremental mapping way~\cite{Maggio2024Clio, werby23hovsg}. On the contrary, we intuitively abstract the neural representations of space to build topo-field in three key aspects: 1) The cognitive map corresponds to a topometric map, which uses graph-like representations to encode relationships among its components, e.g. layouts and objects. 2) The population of place cells is analogous to a neural implicit representation with position encoding, enabling location-specific responses. 3) POR, which prioritizes spatial layouts over content, aligns with our spatial layout encoding of connected regions.

This work proposes a Topo-Field, integrating the Layout-Object-Position (LOP) association into neural field training and constructing a topometric map based on the learned neural implicit representation for hierarchical robotic scene understanding. 
By inputting RGB-D sequences, objects and background contexts are encoded separately as contents and layout information to train a neural field, forming a detailed scene representation. A contrast loss against features from LFMs is employed, resulting in little need for annotation. Further, a topometric map is built by querying the learned field, which is efficient for navigable path planning. To validate the effectiveness of Topo-Field, we conduct quantitative and qualitative experiments on several multi-room apartment scenes evaluating the abilities including position attributes inference, text/image query localization, and planning.

Our contributions can be listed as follows:

\begin{itemize}
\item \textbf{Brain-inspired Topo-Field:} We introduce a Topo-Field that combines neural scene representation with efficient topometric mapping, enabling hierarchical robotic scene understanding and navigable path planning.
\item \textbf{Cognitive Map Representation:} Inspired by the population code in postrhinal cortex (POR) strongly tuned to spatial layouts over scene content rapidly forming a high-level cognitive map, we incorporate the concepts of neural representations of spatial layouts, objects, and place cells to construct hierarchical robotic topometric maps.
\item \textbf{Layout-Object-Position (LOP) Representation:} We develop an implicit neural representation associating layout, object, and position information, which is explicitly supervised using an LFM-powered strategy, requiring minimal human annotation.
\item \textbf{Topometric Map Construction:} We propose a two-stage pipeline for building a topometric map by querying the learned neural field and validating edges among vertices using LLMs, enabling efficient path planning.
\end{itemize}


\section{Related Works}

\subsection{Dense Representation with Neural Radiance Field}
Detailed 3D scene reconstruction has made great efforts in producing lifelike results, among which NeRF (Neural Radiance Fields)~\cite{nerf} has widely attracted researchers' attention. 
A popular research direction is to integrate semantics with NeRF to achieve a more comprehensive understanding of scenes~\cite{snerf-1,snerf-2,snerf-3}. 
Recently, several robotic works have demonstrated that features from LFMs can be used for self-supervised learning, which reduces the costly manual annotation~\cite{clip-fields, vlmaps, lerf2023}. 
However, they focus on object semantics but do not include layout-level features. 
RegionPLC~\cite{yang2023regionplc} considered region information but with no explicit representation of layout features.
In contrast, in our work, CLIP~\cite{clip} and Sentence-BERT~\cite{sentence-bert} are employed to generate vision-language and semantic features for objects and layout learned respectively. 

\subsection{Topometric Map for Scene Structure Understanding}
Using detailed neural fields for planning and navigation is computationally demanding, on the other hand, hybrid topometric mapping has been known for its efficiency in terms of managing the information and being queried for downstream tasks~\cite{topo1, topo2, topo3}. It takes advantage of both metric maps and topological maps. 
However, most topological maps have not introduced information such as semantics. This makes it unsuitable for language/image-guided planning tasks, which is a growing trend in scene representation applications. 
Concept-graph~\cite{gu2024conceptgraphs} makes a step forward utilizing LFM to model the object structure with a topo map. 
CLIO~\cite{Maggio2024Clio}built a task-driven scene graph forming task-relevant clusters of primitives. HOV-SG~\cite{werby23hovsg} proposed using feature point cloud clustering and mapping in an incremental approach. On the contrary, we propose to build the topometric map by querying the trained neural field which serves as knowledge-like memory base, whose nodes and edges include attributes representing object and layout information.

\subsection{Spatial Understanding with Layout Information}
Generally, topology is built based on clustering from occupancy information or Voronoi diagrams~\cite{he2021hierarchical}, regardless of the contents and layout relationship. However, neuroscience findings suggest a mechanism to form a high-level cognitive map from egocentric perception to allocentric representation~\cite{zeng2022theory, tolman1948cognitive}.
Place cells~\cite{o1971hippocampus}, as the embodiment of cognitive map, together with spatial view cells show activity to contents~\cite{rolls1998information}. Recently, Patrick et al.~\cite{por-cells} showed that a population code in the POR is more strongly tuned to the spatial layout than to the content in a scene. This suggests that there are specialized cells and signaling mechanisms to process layout in the process of scene understanding, which captures the spatial layout of complex environments to rapidly form a high-level cognitive map representation~\cite{zeng2022theory}.
Inspired by the above research, we mimic the neural scene understanding mechanism by employing egocentric neural field with content and layout knowledge to construct allocentric topometric map.



\section{Overview} 
\label{overview}

\begin{figure*}[t]
\centering
\includegraphics[width=0.85\linewidth]{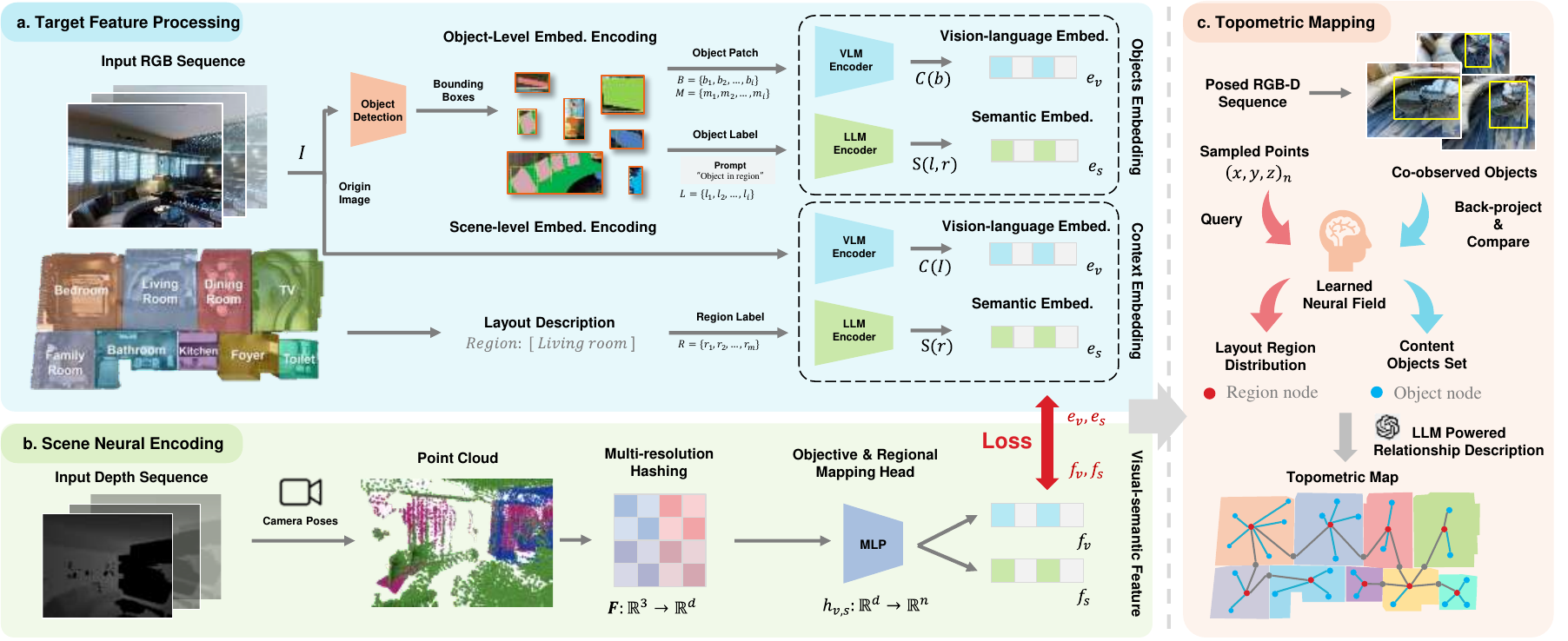}
\vspace{-0.05in}
\caption{\textbf{Pipeline of the Topo-Field.} \textbf{(a)} The ground truth generation of layout-object-position vision-language and semantic embeddings for weakly-supervising. \textbf{(b)} The neural implicit network mapping 3D positions to target feature space. A contrastive loss is optimized against each other. \textbf{(c)} Topometric mapping process with trained neural field.}
\label{fig:method}
\vspace{-0.05in}
\end{figure*}

We propose to learn an implicit representation of a scene with the neural encoding approach by establishing associations between 3D positions and their corresponding layout and object features as the scene knowledge. Then, a topometric map is built with the learned neural field to form an efficient and queriable representation with a comprehensive understanding of the scene. 
Therefore, we need to train a scene-dependent implicit function, denoted as 
\begin{equation}
F:\mathbb{R}^3 \to \mathbb{R}^n,
\end{equation}
where for any 3D point $P$ in space, $F(P)$ is supposed to match with $\mathcal{E}\{(e_v, e_s)\} \in \mathbb{R}^n,$ representing the layout-object-position associated embedding of that point $e_v$ and $e_s$ are vision-language embedding and semantic embedding of image point where $P$ is back-projected from.
CLIP~\cite{clip} image encoder is introduced to encode $e_v$ integrating the vision and language feature space. Besides, the Sentence-BERT~\cite{sentence-bert} feature is also introduced to encode $e_s$ in this work. Because intuitively, unlike objects that can have similar appearances within a certain category, region information often lacks specific visual appearances and is closely related to semantic representations like the integration purpose of the scene and object semantics. Models trained on large-scale question-answering datasets can aid in understanding the semantic relationships between regions and objects. 
Target feature processing and training strategy to match the embeddings to targets are described in Section \ref{method_feature_processing} and \ref{method_training}. Applications utilizing the learned field are discussed in Section \ref{method_association}. 

Based on the trained $F$, we aim to build a topometric map denoted as 
\begin{equation}
G = (V, E),
\end{equation}
where vertices $V$ include object vertices $\mathbf{v}_o$ and region vertices $\mathbf{v}_r$ and
edges $E$ include edges between objects $\mathbf{e}_{o-o}$, edges between regions $\mathbf{e}_{r-r}$, and edges between object and region $\mathbf{e}_{o-r}$. The topological map architecture and construction pipeline are described in Section \ref{method_topomap}.
\section{Method}\label{method}

\begin{figure*}[t]
\centering
\includegraphics[width=0.8\linewidth]{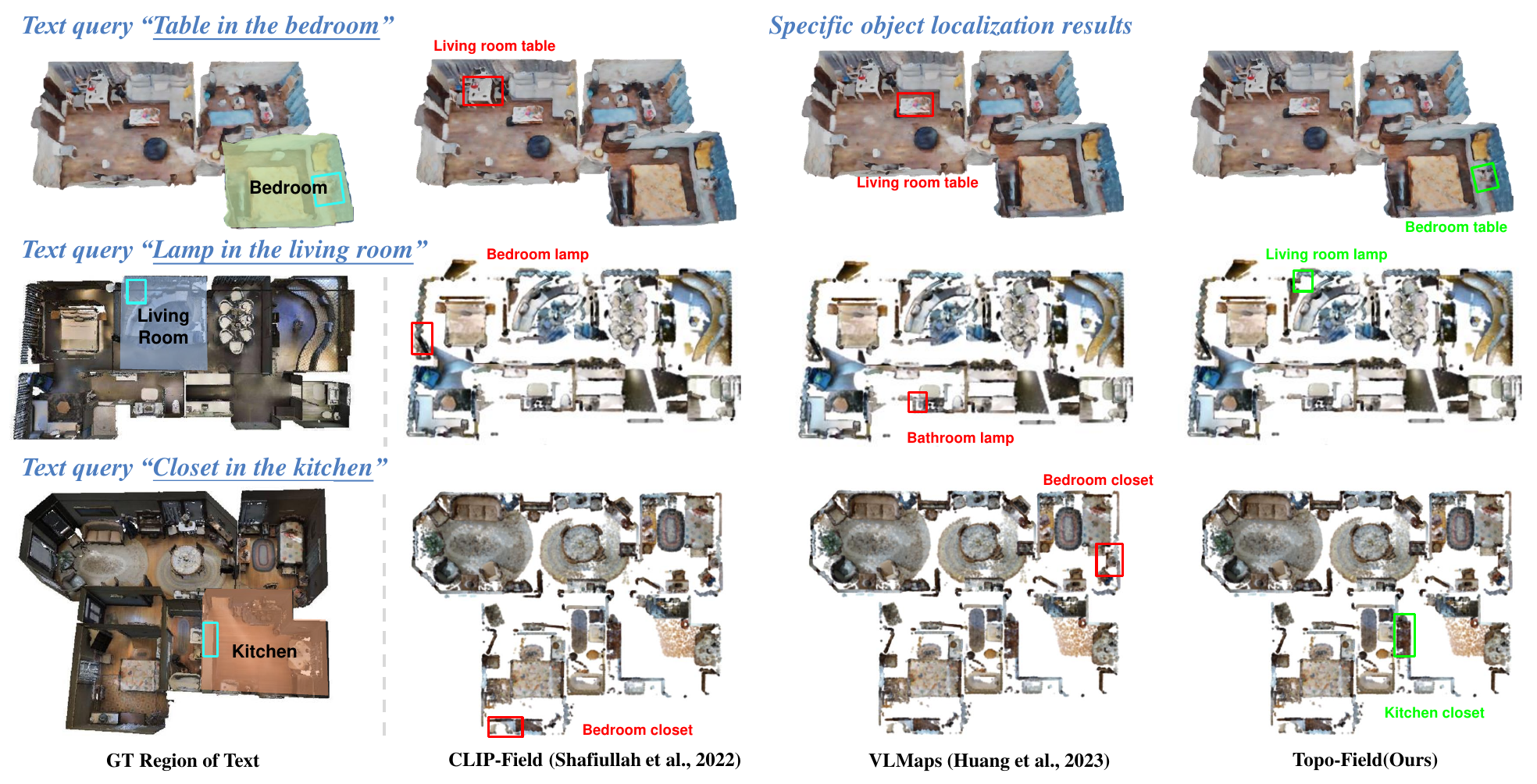}
\vspace{-0.1in}
\caption{\textbf{Qualitative comparison of text query localization} results among state-of-the-art methods and our method with text input in the form of ``\textit{object in the region}''. Blue box shows the ground truth bounding box of object. Red box means miss-predicted box, while green box means the correctly predicted results.}
\label{fig:exp3}
\end{figure*}

\subsection{Target Feature Processing}
\label{method_feature_processing}
RGB-D image sequences with poses are accepted as input to get the target layout-object-position features for training $F$. For pure RGB image sequences, depth point clouds and camera poses can also be estimated through methods like COLMAP~\cite{colmap} or simultaneous localization and mapping (SLAM).
The only employed GT annotation is the layout distribution of environment where the region of each 3D point $P$ is denoted as $r_P \in R=\{r_1, r_2, \dots, r_q\}$, where $q$ is the number of regions. Such information is available in datasets like Matterport3D~\cite{Matterport3D}. 
However, in fact, partitioning the buildings needs little human labor, where in most human-made buildings spatial layouts are easily available divided by straight walls. As in our practice, region annotation of a house with 8 rooms only takes 3 min by drawing lines from top-down view according to walls to form a rule to separate $(x,y)$ coordinates, bounding 3D points to different regions.

For each image $I$, we employ Detic~\cite{detic} $D$ to generate object instance patches with number $i$, including bounding-boxes $B=\{b_1,b_2,\dots ,b_i\}$, masks $M=\{m_1,m_2,\dots ,m_i\}$, and labels $L=\{l_1, l_2, \dots ,l_i\}$.

For object pixels $p_o$ in instance mask $j$, CLIP~\cite{clip} $C$ is employed to compute per-pixel features in mask $b_j$ and Sentence-BERT~\cite{sentence-bert} $S$ is employed to process the semantic feature of $l_j$, prompted in the form of ``$l_j$ \textit{in} $r_{p_o}$''. Given the related region $r_{p_o}$ of $p_o$, embedding of $p_o$ can be denoted as $ e_{p_o} = \{C(b_j), S(l_j,r_{p_o})\}$.

What's more, the background appearance is also considered which we proposed to include context information for region layout.
For background pixels $p_b$ out of masks, per-pixel feature of the whole image $I$ is encoded. Its related region $r_{p_b}\in R=\{r_1, r_2, \dots, r_m\}$ is regarded as the text label and embedding of $p_b$ can be calculated as $e_{p_b} = \{C(I), S(r_{p_b})\}$.

Then, pixel-wise embeddings are back-projected to 3D space based on depth and pose and averagely counted to form a distilled 3D feature point cloud. Consequently, the target feature space $\mathcal{E}\{(e_v, e_s)\}$ consists of object and layout features, where $(e_v, e_s)$ directs from
$\{e_{p_o}, e_{p_b}\}_{p_o,p_b \in P}$.
The pipeline is shown in Fig. \ref{fig:method}

Compared with previous implicit neural field methods, $(e_v, e_s)$ includes (1) separately encoded vision-language and semantic information by supervising embeddings from object and background pixels. (2) region information consisted of vision-language embeddings from per-pixel image encoding and semantic embeddings from region text labels. (3) context included object label in the form of ``$l_p$ \textit{in} $r_p$'', where $l_p$ and $r_p$ is object label and region label at point $p$ (e.g., cup in the kitchen). Ablation studies of these improvements are conducted in Section \ref{exp_ablation} with more details.

\begin{figure}[t]
\centering
\includegraphics[width=\linewidth]{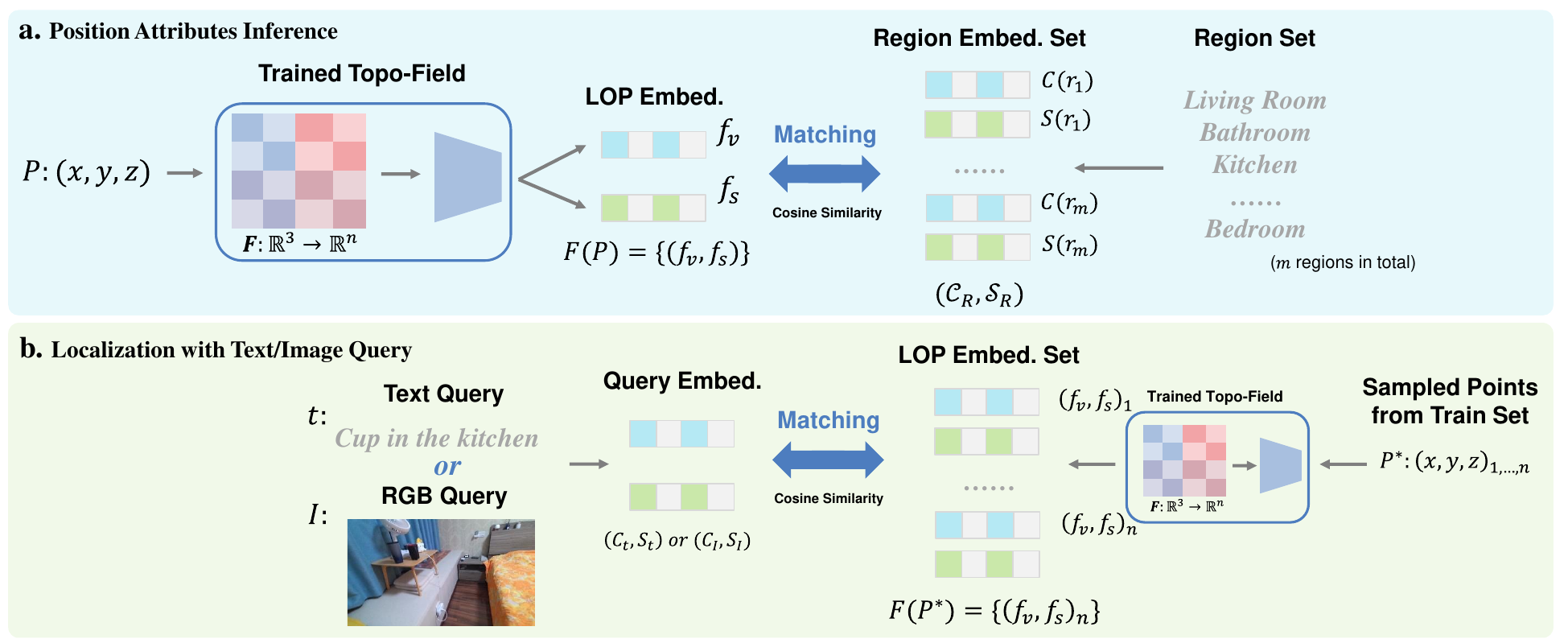}
\vspace{-0.15in}
\caption{\textbf{Capabilities of the learned neural field}. \textbf{(a)} The attributes inference using position input. \textbf{(b)} The LOP association helped localization of text and image queries.}
\label{fig:infer}
\end{figure}

\subsection{Scene Neural Encoding}
\label{method_architecture}

Our proposed Topo-Field involves an implicit mapping function to encode the 3D position into a spatial vector representation $g: \mathbb{R}^3 \to \mathbb{R}^d$ and separate heads $h: \mathbb{R}^d \to \mathbb{R}^n$ processing encodings to match the target feature space $\mathcal{E}\{(e_v, e_s)\}$. 
To select an appropriate implicit function, considering that the target feature space includes object-level local features and layout-level region feature representation, we employ the Multi-scale Hash Encoding (MHE) introduced in Instant-NGP~\cite{instantngp} as $g$ with $d=144$. The feature pyramid structure used in MHE allows for considering structural features ranging from coarse to fine in the spatial domain. Additionally, MHE has a faster training speed compared to traditional NeRF~\cite{nerf} network structures.
For mapping the position encodings to the target feature space, we employ a unified and simple Multi-Layer Perceptron (MLP) network structure. It includes heads $h_v: \mathbb{R}^d \to f_v$ for obtaining vision-language features and $h_s: \mathbb{R}^d \to f_s$ for semantic features, which together form the high dimension embeddings $\{f_v, f_s\}\in \mathbb{R}^n$. The model is shown in Fig. \ref{fig:method}.

In this way, given a posed RGB-D image, the target feature of each pixel is processed as mentioned in Section \ref{method_feature_processing} denoted as $\mathcal{E}\{(e_v, e_s)\}$. At the same time the related pixel in depth image is back-projected into 3D space according to depth and pose value and processed by the above mentioned $g, h$ to form $\{f_v, f_s\}$. A contrastive loss is conducted between $\{(e_v, e_s)\}$ and $\{f_v, f_s\}$ to train the neural representation. Training details are declared in Section \ref{method_training}.

\begin{table}[tp]
\footnotesize
\centering
\setlength{\abovecaptionskip}{0.1cm}
\renewcommand{\arraystretch}{1.1}
\setlength{\tabcolsep}{3.2pt}
\begin{tabular}{lcccccccc}
\toprule 
\multirow{2}{*}{\textbf{Methods}} & \multicolumn{2}{c}{\textbf{Scene1}} & \multicolumn{2}{c}{\textbf{Scene2}} & \multicolumn{2}{c}{\textbf{Scene3}} & \multicolumn{2}{c}{\textbf{Scene4}} \\ \cline{2-9}
                      & Dist. & Acc. & Dist. & Acc. & Dist. & Acc. & Dist. & Acc. \\ \hline
CLIP-Field(2022)            & 2.97 & 0.24 & 3.35 & 0.21 & 2.98 & 0.20 & 3.06 & 0.17 \\
VLMaps(2023)                & 2.78 & 0.28 & 3.63 & 0.16 & 3.05 & 0.24 & 3.12 & 0.12 \\
LERF(2023)                  & 2.86 & 0.32 & 2.82 & 0.11 & 3.49 & 0.17 & 3.04 & 0.20 \\
Topo-Field             & \textbf{0.92} & \textbf{0.85} & \textbf{0.86} & \textbf{0.84} & \textbf{0.36} & \textbf{0.95} & \textbf{0.27} & \textbf{0.97} \\
\midrule 
\midrule 
Text queries               & \multicolumn{2}{c}{100} & \multicolumn{2}{c}{100} & \multicolumn{2}{c}{60} & \multicolumn{2}{c}{60} \\
\bottomrule 
\end{tabular}
\caption{\textbf{Quantitative comparison of text query localization} results on different scenes from the Matterport3D dataset. The average distance (m) from the target to the localized point cloud and the accuracy evaluating whether predicted positions are in the correct region are used as metrics.}
\label{exp_2_q}
\vspace{-0.2in}
\end{table}

\subsection{Topometric Mapping}
With the function and feature representation mentioned above, we can integrate 3D positions with the object and region information and construct a topometric map. 
The topo map construction process is formed in a mapping and updating strategy, while the implicit neural representation is introduced and queried as scene knowledge in this process. Detailed pipeline is introduced as follows.

\subsubsection{Knowledge from Learned Neural Field}
\label{method_association}

\textbf{Position Attributes Inference.} Using spatial 3D point $P$ as input, assuming a collection of space regions $R$ (e.g., ``living room''``bathroom''``bedroom''\dots), 
we compute the vision-language features $\mathcal{C}_R=\{C(r_1), C(r_2), \dots, C(r_m)\}$ and semantic features $\mathcal{S}_R=\{S(r_1), S(r_2), \dots, S(r_m)\}$ using CLIP~\cite{clip} encoder $C$ and Sentence-BERT~\cite{sentence-bert} encoder $S$, where $m$ is the number of rooms. 
Then the cosine similarity between $F(P)=\{(f_v, f_s)\}_P$ and $\{\mathcal{C}_R, \mathcal{S}_R\}$ is calculated to find the most likely region to which $P$ belongs. The inference process is shown in Fig. \ref{fig:infer} (a).
Similarly, the object information of $P$ can be inferred with the same approach replacing the region set $R$ with object set $O$.

\textbf{Localization with Text/Image Query.} For natural language text input $t$ (e.g., ``cup in the bedroom"), most existing robotic scene representations struggle to locate specific objects of interest (e.g., differentiating between cups in the living room and the bedroom). However, with our proposed Topo-Field that includes region information, 
we can calculate the cosine similarity between $\{\mathcal{C}_t, \mathcal{S}_t\}$ and the embeddings $F(P^*)=\{(f_v, f_s)\}_{P^*}$ to find the most likely position of queries, where $P^*$ are sampled from 3D points set to train $F$. As for image input $I$, we can calculate the cosine similarity of $\{\mathcal{C}_I, \mathcal{S}_I\}$ with $F(P^*)=\{(f_v, f_s)\}_{P^*}$ in the same way to find the 3D points set with highest similarity.
Localization process of text query and image query is shown in Fig. \ref{fig:infer}.

\begin{figure*}[t]
\centering
\includegraphics[width=0.8\linewidth]{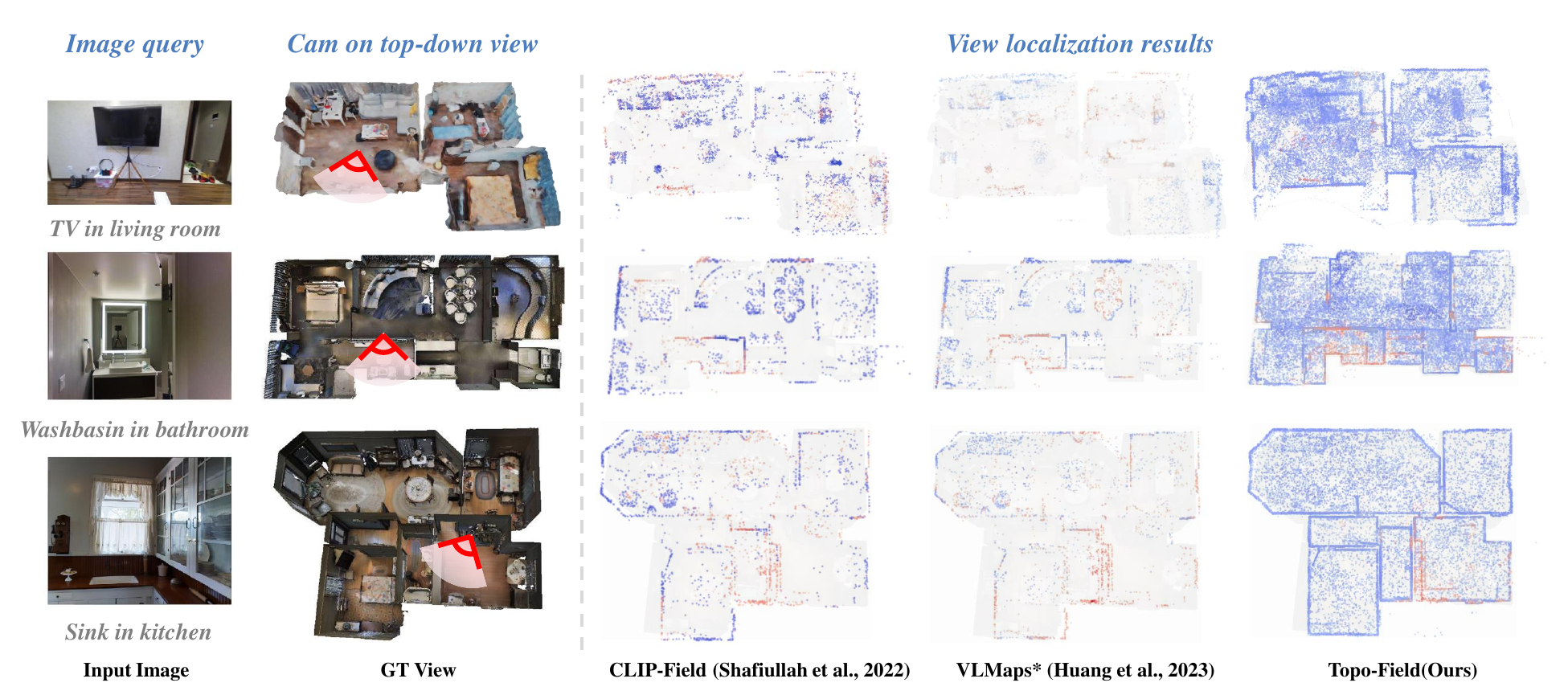}
\caption{\textbf{Qualitative comparison of image query localization} results in heatmaps form among state-of-the-art methods and our method with image input. Our approach localizes the position of queried image in an exact smaller range.}
\label{fig:exp4}
\vspace{-0.05in}
\end{figure*}

\subsubsection{Topometric Map Construction}
\label{method_topomap}

As defined in Section \ref{overview}, topometric map $G=(V, E)$ consists of vertices and edges. We define a vertice $\mathbf{v}$ \{ id, node\_type, class, bounding\_box, caption\} and edge $\mathbf{e}$ \{ id, edge\_type, start\_node, end\_node, relationship, caption \}.
Mimicking the mental representation of cognitive maps, we construct the topometric map in a \textbf{mapping and updating} strategy based on the learned Topo-Field $F$.

\textbf{Mapping}. we first averagely sample $k$ points $P_{1,\dots ,k}$ in the environment (each grid of $0.5m \times 0.5m$ with a point in our practice) and infer their related regions according to Section \ref{method_association}. Supposing there are $m$ regions in total $r_{1,\dots ,m}$, we calculate the extent of each region in the bounding-box format according to positions of points within the same region. The topo map region vertice set is then initialized as $\mathbf{v}_r=\{\mathbf{v}_{r_1},\mathbf{v}_{r_2},\dots ,\mathbf{v}_{r_m}\}$. For each $\mathbf{v}$, \{id\} is set, \{node\_type\} is \{region\}, \{class\} and \{caption\} is set according to the inferred region label, and \{bounding\_box\} is set to the bound of coordinates.
On the other hand, while employing Detic~\cite{detic} to detect object instances as mentioned in Section \ref{method_feature_processing}, instances with high confidence (more than 60\% in our practice) are recorded as object vertices candidates. For each $\mathbf{v}$, \{node\_type\} is \{object\}, \{class\} and \{caption\} is set according to the prediction result, and \{bounding\_box\} is set according to the back-projected masked pixels similarly.
With the mapped nodes, we leverage LLM to describe the layouts with connectivity, distances, and relationships of regions and objects in JSON format based on the vertices' attributes and poses. During this process, edges are built among vertices. For object-object edge $\mathbf{e}_{o-o}$, we follow ~\cite{gu2024conceptgraphs} which mainly consider bounding-box overlap. For object-region edge $\mathbf{e}_{o-r}$, we consider an object belongs to the region if the object b-box is in the region b-box and filter the unreasonable relation noise powered by LLM (e.g., it's almost impossible that a bike is in bedroom). For region relationships, the adjacency and position relationship of region b-box is considered. Examples of LLM prompts to build relationships and JSONs are listed in appendix for reference. Fig. \ref{fig:method} shows the pipeline of metric-topological map construction.

\textbf{Updating}. 
RGB-D image sequence for training $F$ or a newly captured sequence can be used for constructed topometric map fine-tuning. For object vertices, if an object is detected by more than $3$ frames in sequence, the object b-box will be compared with the constructed vertices. A new vertice will be added if no vertice corresponds to it with the above-mentioned process. For region vertices, we calculate embeddings $F(p_I)$ of sampled back-projected pixels $p_I$ in each image $I$. $F(p_I)$ will be matched with the constructed region set $r_{1,\dots ,m}$, and extent of a region $r$ will be updated if $F(p_I)$ matches $\{\mathcal{C}_r, \mathcal{S}_r\}$ and $p_I$ exceeds the \{bounding\_box\} extent of vertice $\mathbf{v}_r$. LLM to update edges will be called each 50 frames.

\begin{table*}[t]
\centering
\footnotesize
\setlength{\abovecaptionskip}{0.1cm}
\renewcommand{\arraystretch}{1.1}
\setlength{\tabcolsep}{2.5pt}
\begin{tabular}{lcccccccccc}
\toprule 
\textbf{Methods} & \textbf{Scene1} & \textbf{Scene2} & \textbf{Scene3} & \textbf{Scene4} & \textbf{Scene5} & \textbf{Scene6} & \textbf{Scene7} & \textbf{Scene8} & \textbf{Scene9} & \textbf{Scene10} \\ \hline
CLIP-Field(2022)           & 0.242 & 0.165 & 0.130 & 0.142 & 0.127 & 0.138 & 0.227 & 0.200 & 0.102 & 0.060 \\
VLMaps(2023)               & 0.177 & 0.194 & 0.127 & 0.098 & 0.148 & 0.187 & 0.199 & 0.221 & 0.092 & 0.087 \\
LERF(2023)                 & 0.268 & 0.189 & 0.165 & 0.153 & 0.136 & 0.169 & 0.216 & 0.252 & 0.110 & 0.091 \\
RegionPLC(2023)            & 0.290 & 0.202 & 0.173 & 0.168 & 0.152 & 0.154 & 0.243 & 0.248 & 0.086 & 0.088 \\
Topo-Field            & \textbf{0.886} & \textbf{0.900} & \textbf{0.884} & \textbf{0.894} & \textbf{0.872} & \textbf{0.858} & \textbf{0.901} & \textbf{0.897} & \textbf{0.821} & \textbf{0.839} \\ 
\midrule 
\midrule 
Position Samples   & 169k & 185k & 111k & 112k & 106k & 176k & 130k & 121k & 205k & 211k \\
\bottomrule 
\end{tabular}
\caption{\textbf{Comparison of position attributes inference results} on the test set of different scenes from the Matterport3D dataset. The average region prediction accuracy of sampled 3D points is used as metric.}
\label{exp_1}
\vspace{-0.1in}
\end{table*}

\subsection{Training}
\label{method_training}
The pipeline of ground truth data generation is described in Section~\ref{method_feature_processing} to train $F$. To fit the implicit representation introduced in Section \ref{method_architecture} to the target feature space, we design the loss function through a contrastive approach. 
For the vision-language feature optimization, the tempered similarity matrix on point $P$ is denoted as
\begin{equation}
\text{Sim}_v=\tau \{f_v\}_P \{e_v\}_P,
\end{equation}
where $\tau$ is the temperature term, $\{f_v\}_P$ and $\{e_v\}_P$ is the calculated implicit representation feature and target embedding according to $P$. Using cross-entropy loss, the vision-language loss can be calculated as
\begin{equation}
\mathcal{L}_v=-exp(-\text{dist}_P) (H(\text{Sim}_v)+H({\text{Sim}_v}^T)),
\end{equation}
where $\text{dist}_P$ is the distance from $P$ to camera, and $H$ is the cross-entropy function. For the semantic loss, similarity on points $P$ can be calculated as
\begin{equation}
\text{Sim}_s=\tau \{f_s\}_P \{e_s\}_P.
\end{equation}
Similarly, semantic loss can be denoted as
\begin{equation}
\mathcal{L}_s=-\text{conf} (H(\text{Sim}_s)+H({\text{Sim}_s}^T)),
\end{equation}
where $conf$ is the prediction confidence from the detection model.
The total loss is computed by:
\begin{equation}
\mathcal{L}=\mathcal{L}_v + \mathcal{L}_s.
\end{equation}
In our experiments, an NVIDIA RTX3090 GPU is utilized and the batch size is set to $12544$ to maximize the capability of our VRAM. 
As model instances, CLIP with SwinB is employed in Detic~\cite{detic}, CLIP~\cite{clip} encoder is ViT-B/32 and Sentence-BERT~\cite{sentence-bert} encoder is all-mpnet-base-v2.
The MHE has $18$ levels of grids and the dimension of each grid is $8$, with $log_2$ hash map size of $20$ and only 1 hidden MLP layer of size $600$. We train the neural implicit network for $100$ epochs with optimizer $Adam$, employing a decayed learning rate of $1e-4$ and $3e-3$ decay rate. Each epoch contains $3e6$ samples.
Codes and scripts are released in supplementary for reproducibility.

\begin{figure*}[t]
\centering
\includegraphics[width=0.75\linewidth]{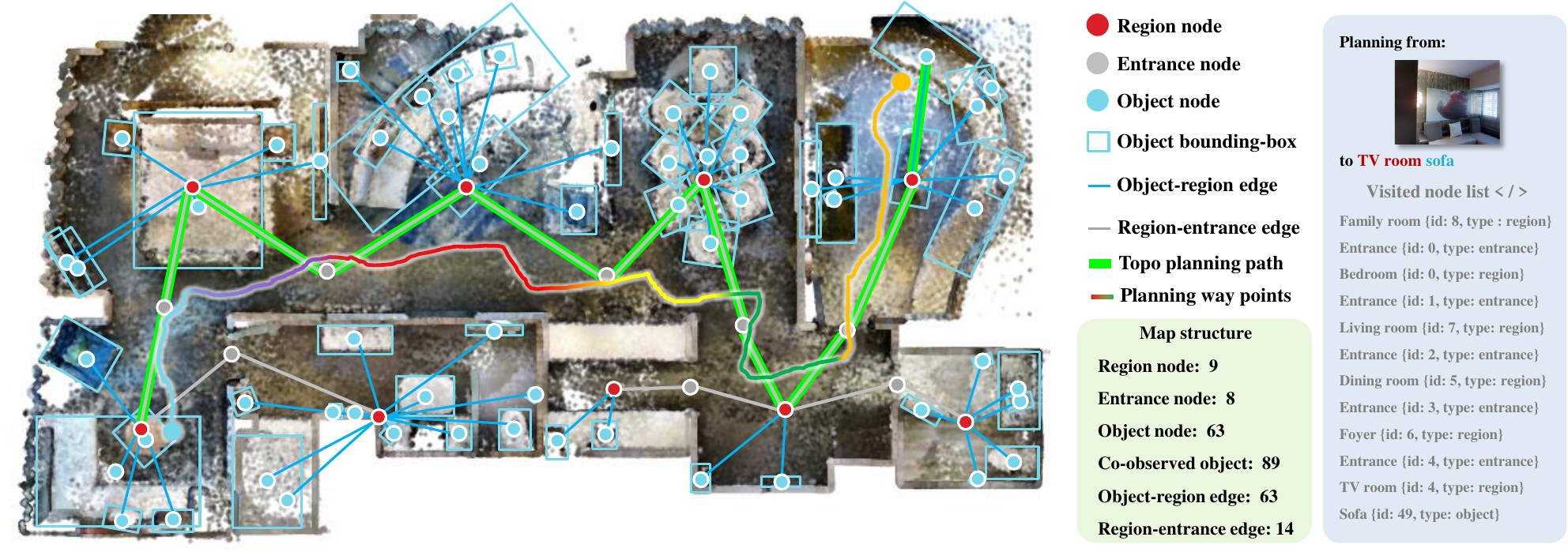}
\vspace{-0.05in}
\caption{\textbf{Topometric map construction example.}
The topometric map is represented as a graph from a top-down view according to the position of nodes. Map structure shows number of nodes and edges. A planning path from a seen view to target is shown as an example employing topometric map, the path is highlighted in green showing the related nodes and edges.  Visited nodes are listed on the right. The line with gradient colors represents the waypoints based on the planning results while different colors represent different predicted regions of waypoints.}
\label{fig:topo-map}
\vspace{-0.1in}
\end{figure*}
\section{Experimental Results}
\label{experiment}

\begin{table}[t]
\centering
\footnotesize
\setlength{\abovecaptionskip}{0.1cm}
\renewcommand{\arraystretch}{1.1}
\setlength{\tabcolsep}{2.5pt}
\begin{tabular}{lcccc}
\toprule 
\textbf{Methods} & \textbf{Scene1} & \textbf{Scene2} & \textbf{Scene3} & \textbf{Scene4} \\ \hline
CLIP-Field(2022)           & 2.541 & 2.748 & 2.922 & 2.651 \\
VLMaps*(2023)               & 2.112 & 1.894 & 1.181 & 1.595 \\
LERF(2023)                 & 1.276 & 1.175 & 1.148 & 1.129 \\
Topo-Field            & \textbf{0.742} & \textbf{0.830} & \textbf{0.374} & \textbf{0.327} \\ 
\bottomrule 
\end{tabular}
\caption{\textbf{Quantitative comparison of image query localization} results with other methods. The similarity weighted average distance (m) between the target view point cloud and the predicted point cloud is evaluated. VLMaps* is a self-implemented version with image localization ability.}
\label{exp_2_image}
\vspace{-0.1in}
\end{table}

Our experiments are conducted on real-world datasets to validate the established layout-object-position association. The data environment is of single-floor residential buildings with multiple rooms which is the common working scenario of household robots widely studied in this field. We employed Matterport3D~\cite{Matterport3D} as well as apartment environment~\cite{niceslam} dataset to demonstrate that our approach can be generalized in diverse scenarios.

\subsection{Position Attributes Inference}
To demonstrate the built LOP association integrates positions with layout features, we designed experiments that accept 3D positions as input to infer the region information. For quantitative evaluation, we divided the RGB-D sequences into training and testing sets. The Topo-Field is trained according to Section \ref{method_training} on the training set and tested in the test set. As the region inference task can be treated as a multi-class classification task for each input, the accuracy, precision, and F1-score are used as metrics. Tab. \ref{exp_1} shows the region inference results on 10 real-world scenes in Matterport 3D~\cite{Matterport3D} with different scales and layouts indicating the average accuracy exceeds 85\%.

\begin{figure}[t]
\centering
\includegraphics[width=1.\linewidth]{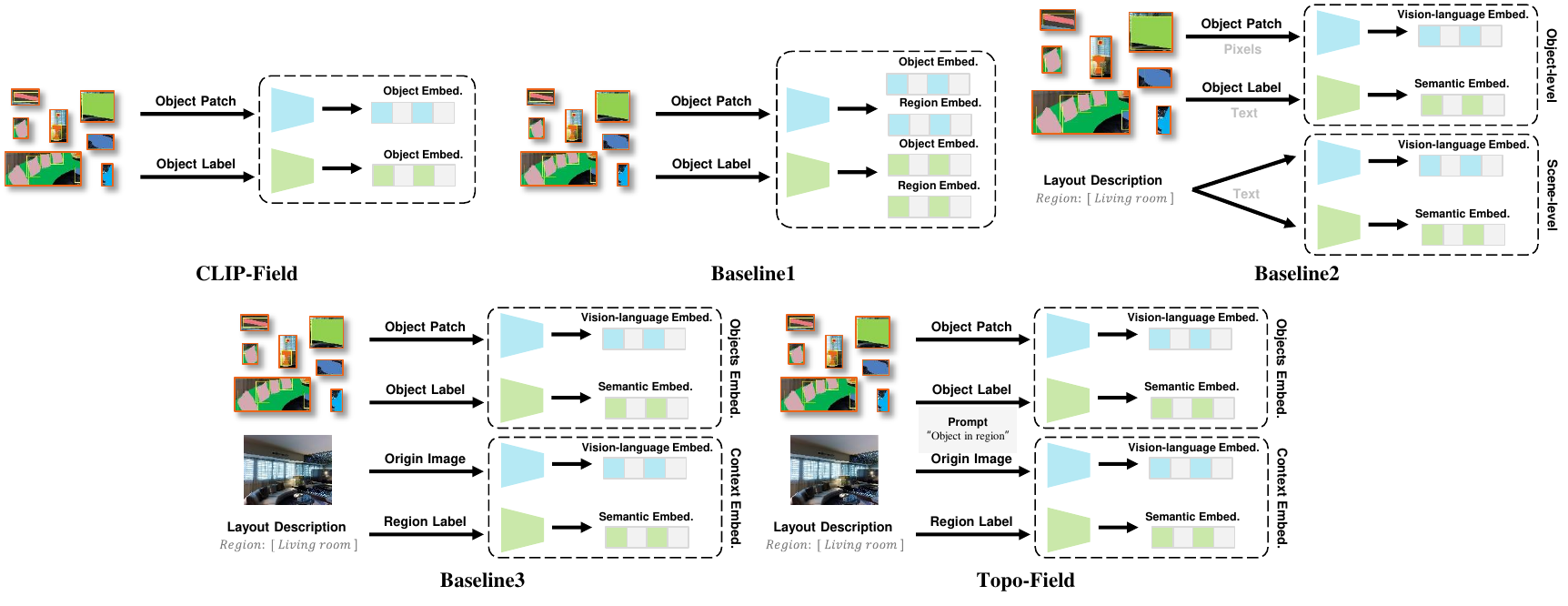}
\vspace{-0.2in}
\caption{Ablation of our LOP information encoding and feature fusion strategy for target features.}
\label{fig::ablation}
\vspace{-0.2in}
\end{figure}

\subsection{Localization with Prompt Queries}
\textbf{Localization with Text Queries:} For objects of the same category in different regions, we input the textual description of the target object in the form of ``object in the region'' and infer the specific location of the target, comparing the results with the predictions from current state-of-the-art visual-language algorithms.
Fig. \ref{fig:exp3} demonstrates the advancements of Topo-Field in object localization tasks involving region information, which allows for the localization of specific target objects based on the description and features of the region, while other methods confuse objects from different regions. Tab. \ref{exp_2_q} shows the quantitative results on 4 scenes of different layouts compared to other methods with an average accuracy of more than $88\%$ and less distance from targets. 
For the metrics, the average distance ($m$) of predicted point cloud and ground truth point cloud is evaluated, together with counting whether the center of predicted points is in the correct room. Ground truth comes from the Matterport3D provided object instance labels.
More results can be seen in the appendix.

\textbf{Localization with Image Queries:} To validate the help of region information in the image view localization task. We localize the images from the test set in the trained Topo-Field. Selected views include representative objects of the scene (e.g., TV in the living room) and views with similar-looking objects or context (e.g., bathroom washbasin and kitchen sink) which is challenging. The localization results are shown in Fig. \ref{fig:exp4} in the form of heatmaps and Tab. \ref{exp_2_image} shows the quantitative results 
which evaluates the weighted average distance of the target view and localized point cloud among all samples in a scene, using similarity as weight.
VLMaps* is a self-implemented version, because origin VLMaps~\cite{vlmaps} does not implement the image localization task. To align with CLIP-Field~\cite{clip-fields} and our work, the LSeg~\cite{lseg} used in VLMap~\cite{vlmaps} is replaced by CLIP~\cite{clip}. The results show that Topo-Field constrains the localization results to a smaller range in the exact region. 
We sampled more than 40 images on each scene from Apartment~\cite{niceslam} and Matterport3D~\cite{Matterport3D} dataset.
By drawing the predicted camera view on the top-down view, we estimated the localization precision and found that most views can be ranged into a specific view on the target field of view, while other methods struggle to get precise results.

\subsection{Topometric Map Construction}

Fig. \ref{fig:topo-map} shows an example of the built topometric map. Layout region nodes, object nodes with bounding boxes, and entrance nodes connecting regions are shown with edges representing relationships. A planned navigable path is shown in the graph from an observed view in family room to the TV room sofa in green. 
The path planning A* algorithm is employed to explore the topological structure to generate waypoints between nodes, and the waypoints are generated with the planning API in Habitat Simulator~\cite{habitat19iccv} and shown in a line with gradient colors, while different colors indicate different predicted regions of the waypoints.

\begin{table}[t]
\centering
\footnotesize
\setlength{\abovecaptionskip}{0.1cm}
\renewcommand{\arraystretch}{1.1}
\setlength{\tabcolsep}{7pt}
\begin{tabular}{ccccc}
\toprule 
\textbf{Methods} & \textbf{Scene1} & \textbf{Scene2} & \textbf{Scene3} & \textbf{Scene4} \\ \hline
CLIP-Field         & 0.242 & 0.165 & 0.130 & 0.142 \\ \hline
Baseline1           & 0.852 & 0.891 & 0.863 & 0.874 \\
Baseline2           & 0.865 & 0.887 & 0.872 & 0.879 \\
Baseline3           & 0.872 & 0.891 & 0.875 & 0.886 \\
Topo-Field           & \textbf{0.886} & \textbf{0.900} & \textbf{0.884} & \textbf{0.894} \\
\bottomrule 
\end{tabular}
\caption{\textbf{Ablation of target feature processing pipeline} of the neural field construction. The average region prediction accuracy of sampled points from different scenes on the Matterport3D dataset is used as the metric.}
\label{exp_ablation}
\vspace{-0.1in}
\end{table}

\subsection{Ablation Study}
Fig. \ref{fig::ablation} and Tab \ref{exp_ablation}. show the ablation of our neural field LOP encoding strategy and feature fusion where:
1) CLIP-Field~\cite{clip-fields} means the origin feature encoding strategy that doesn't explicitly consider the layout features.
2) Baseline1 is our first crude approach that directly supervises the learned embedding from the encoded objects with region semantics.
3) Baseline2 encodes the region description to the target vision-language and semantic feature space for supervision.
4) Baseline3 takes the background pixels into account with the region labels.
5) Topo-Field further considers the context of the layout when supervising the object label semantics.
These four main versions of our numerous iterations of trying are listed as examples to show our work on the neural field encoding of LOP association.

\section{Conclusion and Limitations}\label{conclusion}
We propose a brain-inspired Topo-Field, which integrates Layout-Object-Position (LOP) associations into a neural field and constructs a topometric map from the learned field for hierarchical robotic scene understanding.
However, there are some limitations: 1) Querying and path planning are currently implemented using traditional methods (e.g. A*). Future work will explore using LLMs for more advanced path planning. 2) Real-world deployment on mobile robots for long-term navigation is needed. 3) Future research will focus on updating and editing the topometric map to accommodate environmental changes.

\addtolength{\textheight}{-12cm}   








\bibliographystyle{IEEEtran.bst}
\bibliography{IEEEexample}

\end{document}